\documentclass[]{llncs}
\usepackage{graphicx}
\usepackage{cite} 
\usepackage{times}
\usepackage{epsfig}
\usepackage{amsmath}
\usepackage{amssymb}
\usepackage{mwe}
\usepackage{acro}
\usepackage{amssymb}
\usepackage{xcolor,colortbl}
\usepackage{tabularx}
\usepackage{relsize}
\usepackage{pifont}
\usepackage{booktabs} 
\usepackage{multirow}
\usepackage{multicol}
\usepackage{adjustbox}
\usepackage{float}

\DeclareAcronym{ROI}{
short=ROI,
long=region of interest,
}

\DeclareAcronym{IOU}{
short=IOU,
long=intersection over union,
}

\DeclareAcronym{cIOU}{
short=cIOU,
long=circle intersection over union,
}

\DeclareAcronym{DoF}{
short=DoF,
long=degrees of freedom,
}

\newcommand{\etal}{\textit{et al}. }

\begin{document}
\title{CircleNet: Anchor-free Detection with \\Circle Representation}
%
%
\author{Haichun Yang\inst{1} \and
Ruining Deng \inst{2} \and
Yuzhe Lu \inst{2} \and
Zheyu Zhu \inst{2} \and
Ye Chen \inst{2} \and
Joseph T. Roland \inst{1} \and
Le Lu \inst{3} \and
Bennett A. Landman\inst{2} \and
Agnes B. Fogo\inst{1} \and
Yuankai Huo\inst{2}}
%
\institute{Vanderbilt University Medical Center, Nashville TN 37215, USA \and
Vanderbilt University, Nashville TN 37215, USA \and
PAII Inc., Bethesda MD 20817, USA 
}
%
\maketitle              
\begin{abstract}
Object detection networks are powerful in computer vision, but not necessarily optimized for biomedical object detection. In this work, we propose CircleNet, a simple anchor-free detection method with circle representation for detection of the ball-shaped glomerulus. Different from the traditional bounding box based detection method, the bounding circle  (1) reduces the degrees of freedom of detection representation, (2) is naturally rotation invariant, (3) and optimized for ball-shaped objects. The key innovation to enable this representation is the anchor-free framework with the circle detection head. We evaluate CircleNet in the context of detection of glomerulus. CircleNet increases average precision of the glomerulus detection from 0.598 to 0.647. Another key advantage is that CircleNet achieves better rotation consistency compared with bounding box representations.


\end{abstract}
\section{Introduction}
Detection of Glomeruli is a fundamental task for efficient diagnosis and quantitative evaluations in renal pathology. Recently, deep learning techniques have played important roles in renal pathology to reduce the clinical working load of pathologists and enable the large-scale population based research~\cite{gadermayr2017cnn,bueno2020glomerulosclerosis,govind2018glomerular,kannan2019segmentation,ginley2019computational}. Many traditional feature-based image processing methods have been proposed for detection of glomeruli. Such methods strongly rely on ``hand-crafted" features from feature engineering, such as edge detection~\cite{ma2009glomerulus}, Histogram of Gradients (HOG)~\cite{kakimoto2014automated,kakimoto2015quantitative,kato2015segmental}, median filter~\cite{kotyk2016measurement}, shape features~\cite{maree2016approach}, Gabor filtering~\cite{ginley2017unsupervised}, and Hessian based Difference of Gaussians (HDoG)\cite{zhang2015novel}. 

\begin{figure}
\begin{center}
\includegraphics[width=1\linewidth]{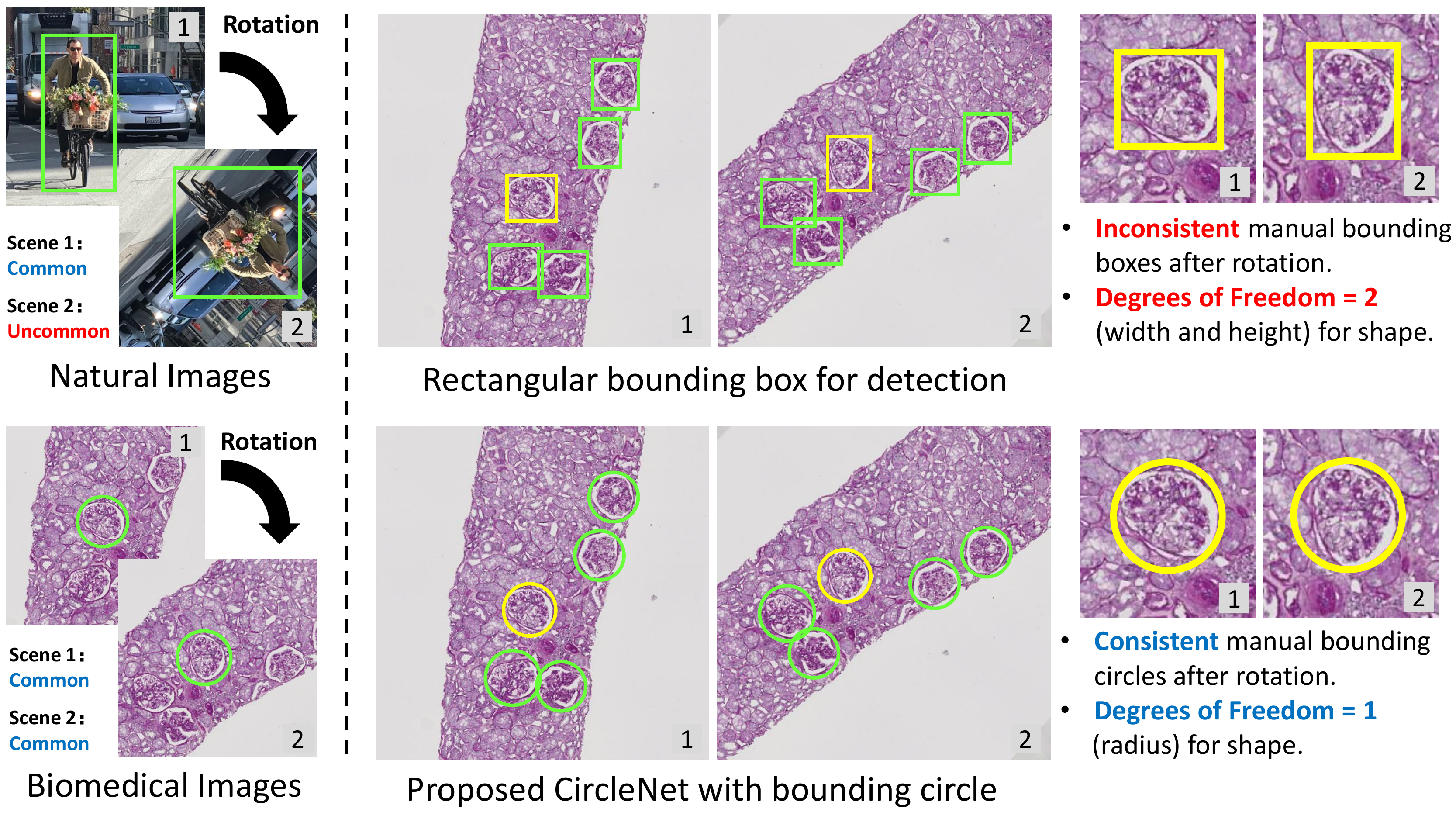}
\end{center}
   \caption{\textbf{Comparison of rectangular bounding box and CircleNet.} The left panel shows that the samples of glomeruli can be acquired and presented with any angles of rotation (both scenes are ``common" for radiologist). The right panel shows that the rectangular bounding box is not optimized for the ball-shaped glomerulus. Using the proposed CircleNet, a more consistent representation with less degrees of freedom is able to be achieved.}
\label{fig:idea}
\end{figure}

In recent years, deep convolutional neural network (CNN) based methods have shown superior performance on detection of glomeruli with ``data-driven" features. Temerinac-Ott \etal\cite{temerinac2017detection} proposed a glomerulus detection method by comparing the CNN performance on different stains. Gallego \etal\cite{gallego2018glomerulus} conducted glomerulus detection by integrating detection and classification, while other researchers \cite{gadermayr2017cnn,bueno2020glomerulosclerosis,govind2018glomerular,kannan2019segmentation,ginley2019computational} combined detection and segmentation. With the rapid development of detection technologies in computer vision, the anchor-based detection methods (e.g. Faster-RCNN~\cite{ren2015faster}) have become the \textit{de facto} standard glomerulus detection approach due to their superior performance. Kawazoe \etal\cite{kawazoe2018faster} and Lo \etal\cite{lo2018glomerulus} proposed the Faster-RCNN based method, which achieved the state-of-the-art performance on glomerulus detection. However, anchor-based methods typically yields higher model complexity and lower flexibility~\cite{ren2015faster,kawazoe2018faster} since anchors are preset on the images and refined several times as detection results. Therefore, recent academic attention has been shifted toward  anchor-free detection methods (means without preset anchors) with simpler network design, less hyper parameters, and even with superior performance~\cite{law2018cornernet,zhou2019objects,zhou2019bottom}.

However, the ``computer vision" oriented detection approaches are not necessarily optimized for biomedical objects, such as the detection of glomeruli, shown in Fig. \ref{fig:idea}. In this paper, we propose a circle representation based anchor-free detection method, called CircleNet, for robust detection of glomeruli. Briefly, the ``bounding circle" is introduced as the detection representation for the ball-shaped structure of the glomerulus. After detecting the center location of the glomerulus, the \ac{DoF} = 1 (radius) are required to fit the bounding circle, while the \ac{DoF} = 2 (height and width) are needed for bounding box. Briefly, the contributions of this study are in three areas:

$\bullet$ \textbf{Optimized Biomedical Object Detection}:  To the best of our knowledge, the proposed CircleNet is the first anchor-free approach for detection of glomeruli with optimized circle representation.

$\bullet$ \textbf{Circle Representation}: We propose a simple circle representation for ball-shaped biomedical object detection with smaller \ac{DoF} of fitting and superior detection performance. We also introduce the \ac{cIOU} and validate the effectiveness of both circle representation and \ac{cIOU}

$\bullet$ \textbf{Rotation Consistency}: The proposed CircleNet achieved better rotation consistency for detection of glomeruli.

\begin{figure}
\begin{center}
\includegraphics[width=0.8\linewidth]{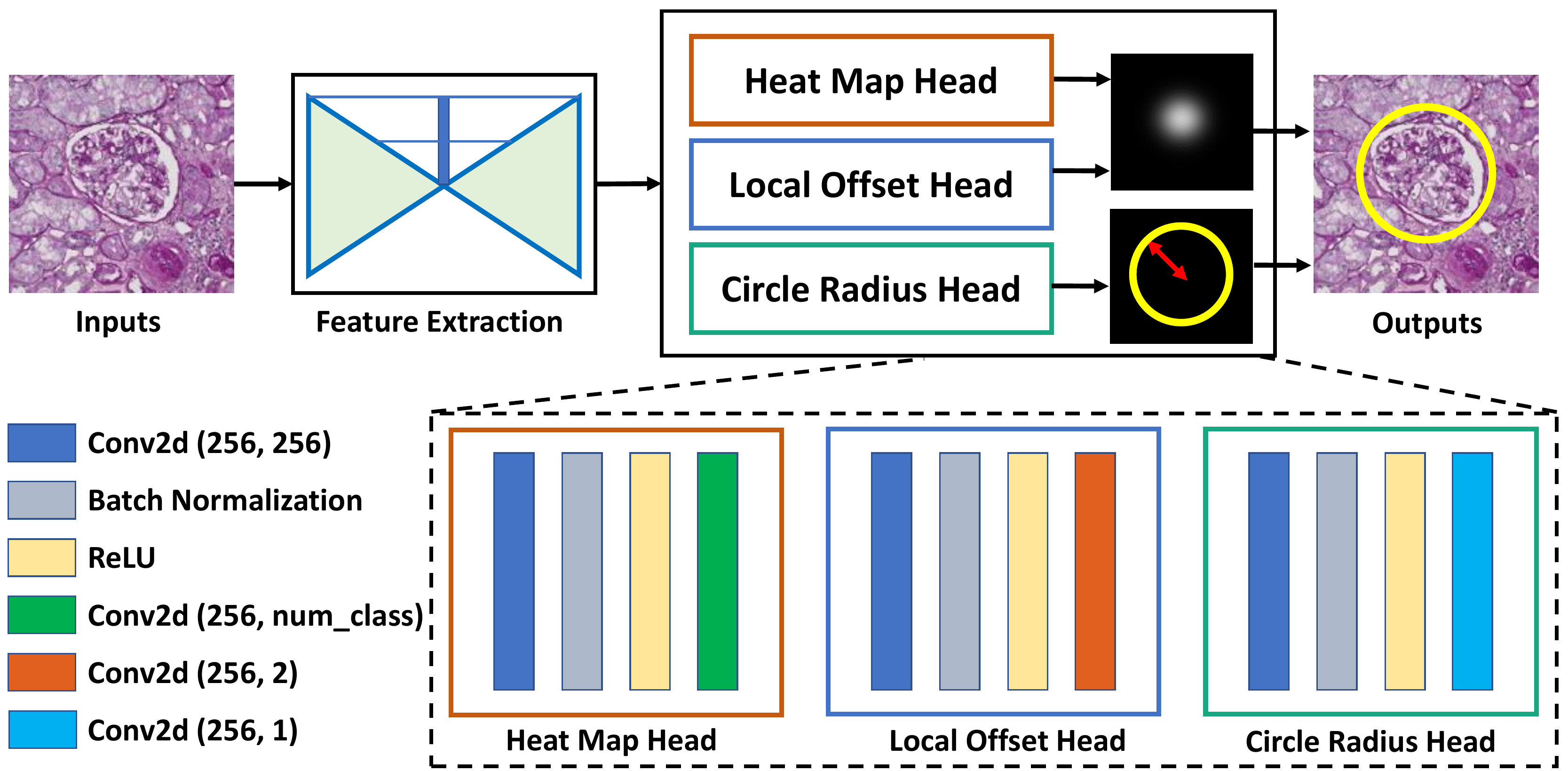}
\end{center}
   \caption{\textbf{The network structure of proposed CircleNet}. The role of the center point localization network is to achieve feature maps for the resulting head networks. Then three network heads are used to achieve the central location and the radius of a bounding circle.}
\label{fig:detection}
 \end{figure}

\section{Methods}
\subsection{Anchor Free Backbone}
In Fig. \ref{fig:detection}, the center point localization (CPL) network is developed based on the anchor-free CenterNet implementation~\cite{zhou2019objects}, as it possesses an ideal combination of high performance and simplicity. Throughout, we follow the definition of the terms from Zhou \etal~\cite{zhou2019objects}. The input image $I$ is defined as $I \in R^{W \times H \times 3}$ with height $H$ and width $W$. The output of the CPL network is the center localization of each object, which is formed as a heatmap $\hat Y \in [0,1]^{\frac{W}{R} \times \frac{H}{R} \times C}$. $C$ indicates the number of candidate classes, while $R$ is the downsampling factor of the prediction. The heatmap $\hat Y$ is expected to be equal to $1$ at lesion centers, while $0$ otherwise. Following standard practice~\cite{law2018cornernet, zhou2019objects}, the ground truth of the target center point is modeled as a 2D Gaussian kernel:
\begin{equation}
{Y_{xyc} = \exp\left(-\frac{(x-\tilde p_x)^2+(y-\tilde p_y)^2}{2\sigma_p^2}\right)}
\end{equation} 
where the $\tilde p_x$ and $\tilde p_y$ are the downsampled target center points and $\sigma_p$ is the kernel standard deviation. The predicted heatmap is optimized by pixel regression loss $L_{k}$ with focal loss~\cite{lin2017focal}:
\begin{equation}
    L_k = \frac{-1}{N} \sum_{xyc}
    \begin{cases}
        (1 - \hat{Y}_{xyc})^{\alpha} 
        \log(\hat{Y}_{xyc}) & \!\text{if}\ Y_{xyc}=1\vspace{2mm}\\
        \begin{array}{c}
        (1-Y_{xyc})^{\beta} 
        (\hat{Y}_{xyc})^{\alpha}\\
        \log(1-\hat{Y}_{xyc})
        \end{array}
        & \!\text{otherwise}
    \end{cases}
\end{equation}
where $\alpha$ and $\beta$ are hyper-parameters in the focal loss~\cite{lin2017focal}. Then, the $\ell_1$-norm offset prediction loss $L_{off}$, is formulated to further refine the prediction location, which is identical as~\cite{zhou2019objects}.

\begin{figure}
\begin{center}
\includegraphics[width=0.6\linewidth]{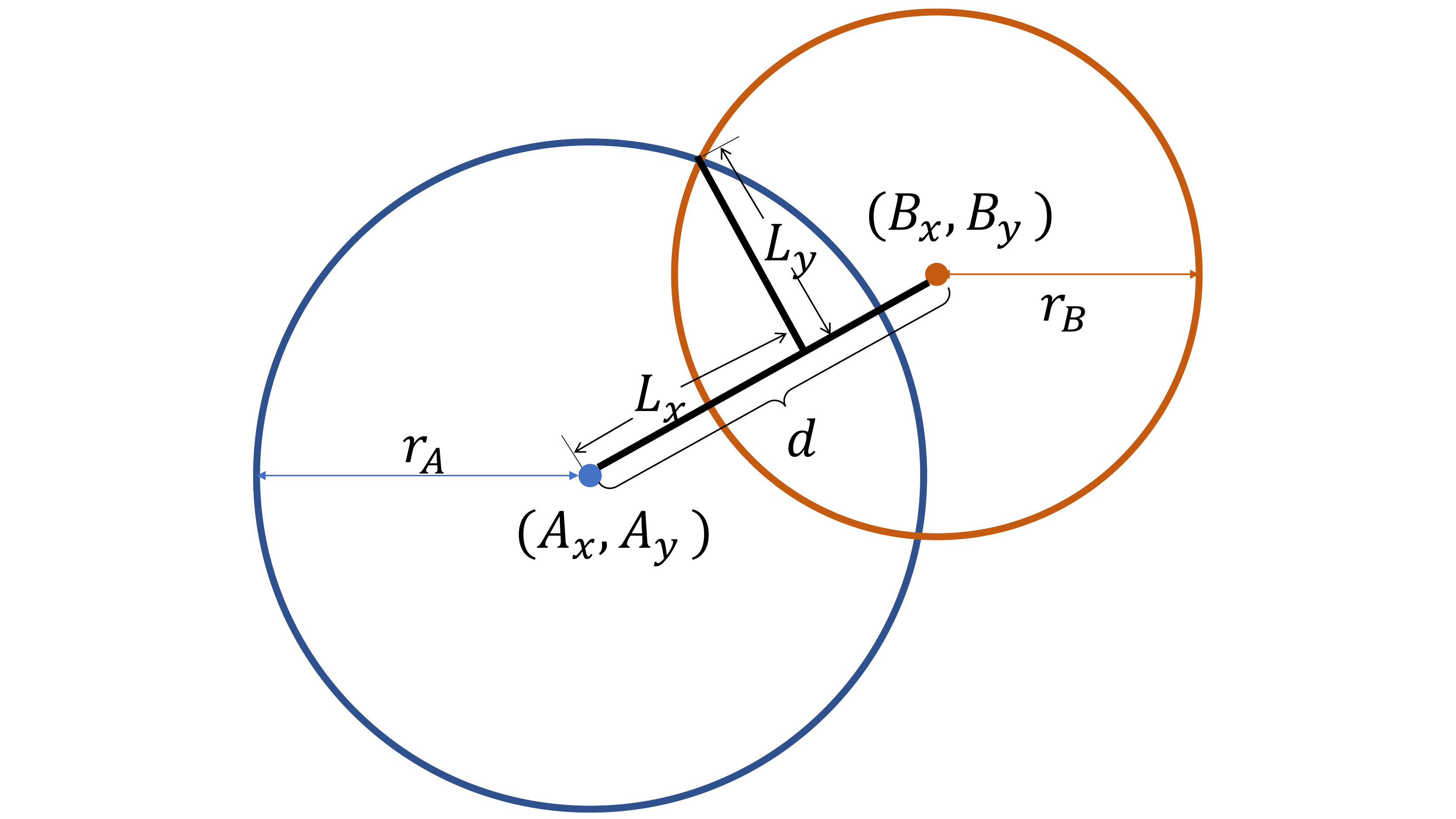}
\end{center}
   \caption{\textbf{The parameters that are used to calculate the circle IOU (cIOU).}}
\label{fig:ciou}
 \end{figure}

\subsection{From Center Point to Bounding Circle}
Once the peaks of the all heatmaps are obtained, the top $n$ peaks are proposed whose value is greater or equal to its 8-connected neighbors.
The set of $n$ detected center points is defined as $\hat{\mathcal{P}} = \{(\hat x_i, \hat y_i)\}_{i = 1}^{n}$.
Each key point location is formed by an integer coordinate $(x_i,y_i)$ from $\hat Y_{x_iy_ic}$ and $L_{k}$. Meanwhile, the offset $(\delta \hat x_i, \delta \hat y_i)$ is obtained from $L_{off}$. Then, the bounding circle is formed as a circle with center point $\hat{p}$ and radius $\hat{r}$ as:
\begin{equation}
 \hat{p} = (\hat x_i + \delta \hat x_i ,\ \ \hat y_i + \delta \hat y_i). \quad \hat{r} = \hat R_{\hat x_i,\hat y_i}.
\end{equation}
where $\hat R  \in \mathcal{R}^{\frac{W}{R} \times \frac{H}{R} \times 1}$ is the radius prediction for each pixel location, optimized by 
\begin{equation}
    L_{radius} = \frac{1}{N}\sum_{k=1}^{N} \left|\hat R_{p_k} - r_k\right|.
\end{equation}
where $r_k$ is the true radius for each circle object $k$. Finally, The overall objective is
\begin{equation}
    L_{det} = L_{k} + \lambda_{radius} L_{radius} + \lambda_{off}L_{off}.
\label{eq:total_loss}
\end{equation} We set $\lambda_{radius} = 0.1$ and $ \lambda_{off} = 1$ referring from ~\cite{zhou2019objects}.

\subsection{Circle IOU}
In canonical object detection, \ac{IOU} is the most popular evaluation metric to measure the similarity between two bounding boxes. The \ac{IOU} is defined as the ratio between area of intersection and area of union. For CircleNet, as the objects are presented as circles, we introduce the \ac{cIOU} as:

\begin{equation}
\textrm{cIOU} = \frac{\textrm{Area} \left( A \cap B \right)}{\textrm{Area} \left( A \cup B \right)}
\end{equation}
where $A$ and $B$ represent the two circles respectively as Fig. \ref{fig:ciou}. The center coordinates of $A$ and $B$ are defined as $(A_x, A_y)$ and $(B_x, B_y)$, which are calculated by:
\begin{equation}
A_x = \hat x_i + \delta \hat x_i, A_y = \hat y_i + \delta \hat y_i
\end{equation}
\begin{equation}
B_x = \hat x_j + \delta \hat x_j, B_y = \hat y_j + \delta \hat y_j
\end{equation}
Then, the distance between the center coordinates $d$ is  defined as:
\begin{equation}
d = \sqrt{\left( B_x - A_x \right)^{2} + \left( B_y - A_y \right)^{2}}
\end{equation}
\begin{equation}
L_x = \frac{r_A^2 - r_B^2+d^2}{2d}, L_y = \sqrt{r_A^2 - L_x^2}
\end{equation}
Finally, the cIOU can be calculated from
\begin{equation}
\textrm{Area} \left( A \cap B \right) = r_A^2\sin^{-1}\left( \frac{L_y}{r_A} \right) + r_B^2\sin^{-1}\left( \frac{L_y}{r_B} \right) - L_y\left(L_x + \sqrt{r_A^2 - r_B^2 +L_x^2}  \right)
\end{equation}
\begin{equation}
\textrm{Area} \left( A \cup B \right) = \pi r_A^2 + \pi r_B^2 - \textrm{Area} \left( A \cap B \right) 
\end{equation}

\section{Data and Implementation Details}

Whole scan images from renal biopsies were utilized for analysis. Kidney tissue was routinely processed, paraffin embedded, and 3$\mu m$ thickness sections cut and stained with hematoxylin and eosin (HE), periodic acid–Schiff (PAS) or Jones. Samples were deidentified, and studies were approved by the Institutional Review Board (IRB). 704 glomeruli from 42 biopsy samples were used as training data, 98 glomeruli from 7 biopsy samples were used as validation data, while 147 glomeruli from 7 biopsy samples were used as testing data. For all training and testing data, the original high-resolution whole scan images (0.25 $\mu m$ per pixel) is downsampled to lower resolution (4 $\mu m$ per pixel), considering the size of a glumerulus~\cite{puelles2011glomerular} and its ratio within a patch. Then, we randomly sampled image patches (each patch contained at least one glomerulus with $512 \times 512$ pixels) as experimental images for detection networks. Eventually, we formed a cohort with 7040 training, 980 validation, and 1470 testing images

The Faster-RCNN~\cite{ren2015faster}, CornerNet~\cite{law2018cornernet}, ExtremeNet~\cite{zhou2019bottom}, CenterNet~\cite{zhou2019objects} were employed as the baseline methods due to their superior performance in object detection. For different detection methods, ResNet-50~\cite{he2016deep}, stacked Hourglass-104~\cite{newell2016stacked} network and deep layer aggregation (DLA) network ~\cite{yu2018deep} were employed as backbone networks, respectively. The implementations of detection and backbone networks followed the authors' official PyTorch implementations. All the models used in this study were initialized by the COCO pretrained model ~\cite{lin2014microsoft}. The same workstation with NVIDIA 1080Ti GPU was used to perform all experiments in this study.

\section{Results}

\begin{table}
\caption{The detection performance.}\label{tab1}
\begin{tabular}{l@{}c@{\ \ }ccccccc}
\toprule
Methods $\quad\quad\quad\quad\quad$ &Backbone & $AP\qquad$ & $AP_{(50)}\quad$ & $Ap_{(75)}\quad$ & $Ap_{(S)}\quad$ & $Ap_{(M)}$ \\
\midrule
Faster-RCNN\cite{ren2015faster} & ResNet-50 & 0.584 & 0.866 & 0.730 & 0.478 & 0.648 \\ 
Faster-RCNN\cite{ren2015faster} & ResNet-101 & 0.568 & 0.867 & 0.694 & 0.460 & 0.633\\
CornerNet\cite{law2018cornernet}& Hourglass-104 & 0.595 & 0.818 & 0.732 & 0.524 & \textbf{0.695} \\
ExtremeNet\cite{zhou2019bottom} & Hourglass-104 & 0.597 & 0.864 & 0.749 & 0.493 & 0.658\\
CenterNet-HG\cite{zhou2019objects} &  Hourglass-104 & 0.574 & 0.853 & 0.708 & 0.442 & 0.649\\
CenterNet-DLA\cite{zhou2019objects} & DLA & 0.598 & 0.902 & 0.735 & 0.513 & 0.648\\
\midrule
CircleNet-HG (Ours) & Hourglass-104  & 0.615 & 0.853 & 0.750 & 0.586 & 0.656 \\ 
CircleNet-DLA (Ours) & DLA & \textbf{0.647} & \textbf{0.907} & \textbf{0.787} & \textbf{0.597} & 0.685  \\ 
\bottomrule
\label{table:detect}
\end{tabular}
\end{table}

\subsection{Detection Performance}
The standard detection metrics were used to evaluate different methods, including average precision ($AP$), $AP_{50}$ (IOU threshold at 0.5), $AP_{75}$ (IOU threshold at 0.75), $AP_S$ (small scale with area $<$ 1000), $AP_M$ (median scale with area $>$ 1000). In the experiment, state-of-the-art anchor-based (Faster-RCNN) and anchor-free detection methods (CornerNet, ExtremeNet, CenterNet) were employed as benchmarks. From Table.\ref{table:detect}, the proposed method outperformed the benchmarks with a remarkable margin, except for $AP_M$. However, the performance of the proposed method still achieved the second best performance for $AP_M$.

\subsection{Circle Representation and cIOU}

\begin{figure}
\begin{center}
\includegraphics[width=0.8\linewidth]{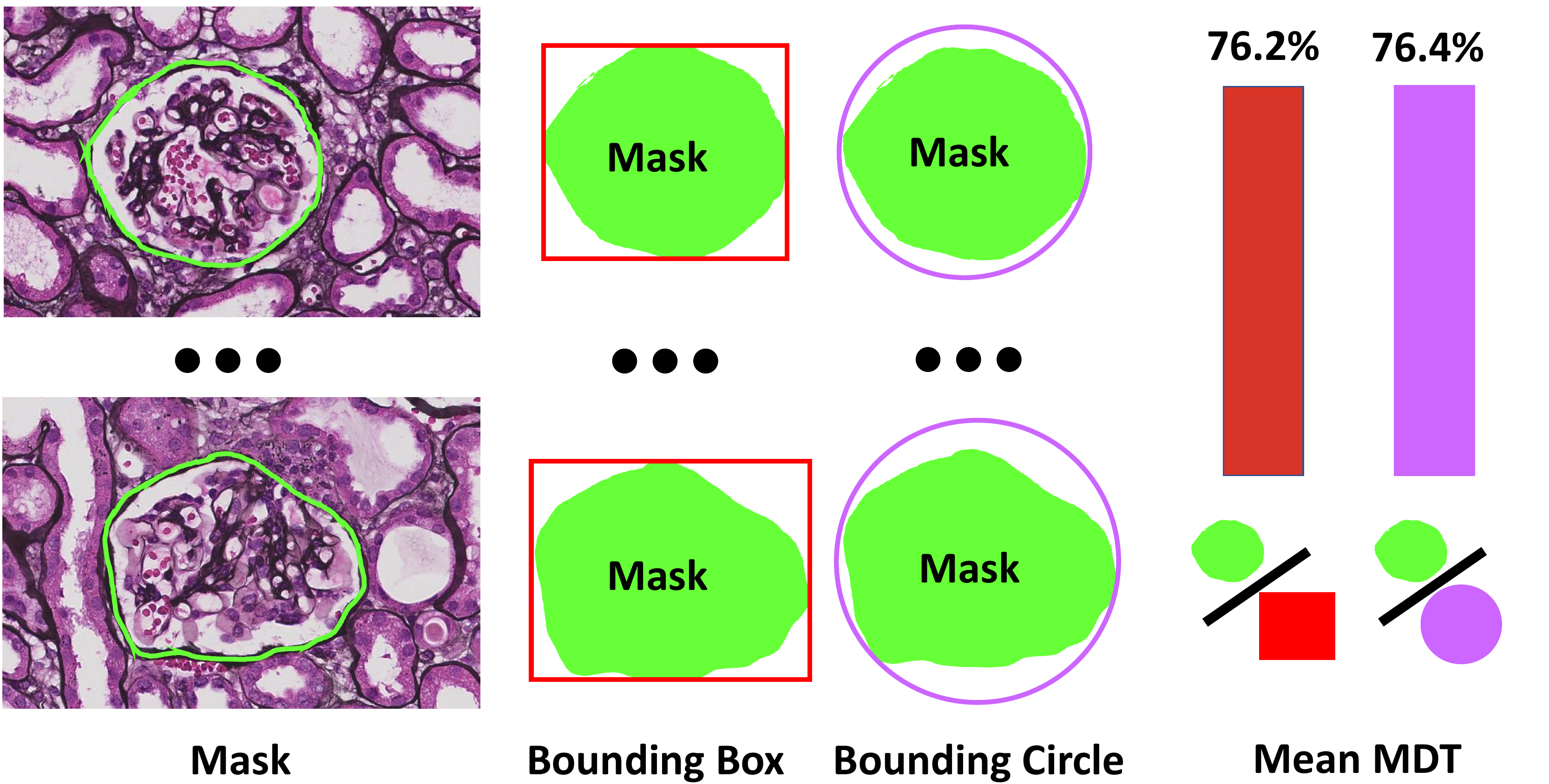}
\end{center}
   \caption{\textbf{The ratio between the mask area and bounding box/circle area, called mask detection ratio (MDT)}. The masks and representations were manually traced (left and middle panel) on 50 randomly selected glomerulus from the testing cohort. As the result, the mean MDT was close across rectangular box representations and circle representations}
\label{fig:mask}
 \end{figure}
 
We further evaluated if the better detection results of circle representation sacrificed the effectiveness for detection representation. To do that, we manually annotated 50 glomerulus from the testing data to achieve segmentation masks. Then, we calculated the ratio between the mask area and bounding box/circle area, called mask detection ratio (MDT), for each glomerulus. From the right panel of Fig. \ref{fig:mask}, both box and circle representations have the comparable mean MDT, which shows that the bounding circle does not sacrifice the effectiveness for detection representation.

\begin{figure}
\begin{center}
\includegraphics[width=0.8\linewidth]{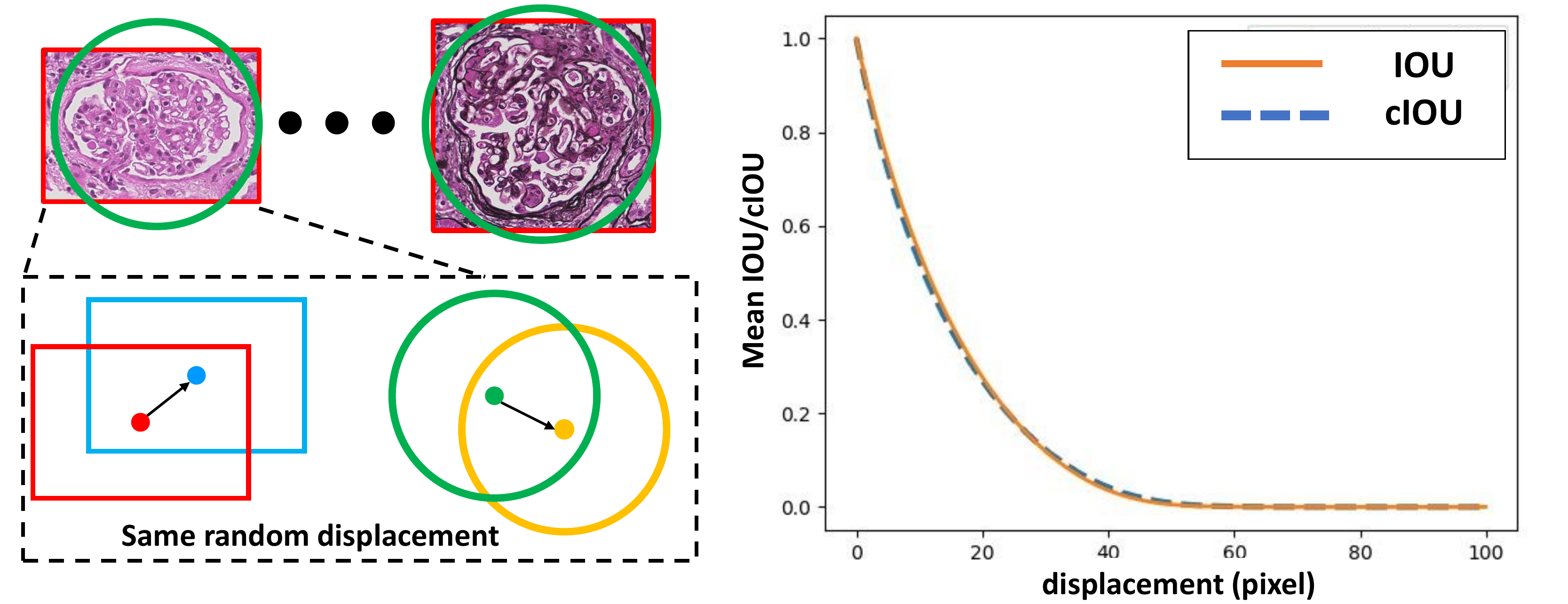}
\vspace{-1em}
\end{center}
   \caption{\textbf{The effectiveness of IOU and cIOU metrics.} For every testing glomerulus, we shifted the bounding box/circle with certain displacement in a random direction (left panel). Then, the IOU and cIOU values between the original and shifted bounding boxes/circles were calculated for different displacement (right panel).}
\label{fig:iou}
 \end{figure}
 
For bounding box and bounding circle, the IOU and \ac{cIOU} were used as the overlap metrics for evaluating the detection performance (e.g., $AP_{50}$ and $AP_{75}$) between manual annotations and predictions. Herein, we compared the performance of bounding box and bounding circle as similarity metrics for detection of glomeruli. To test this, we added random displacements with random directions on all testing glomeruli to simulate different detection results. Then, we calculated the IOU and \ac{cIOU}, respectively. Fig. \ref{fig:iou} presents the results of mean IOU and mean \ac{cIOU} on all testing glomeruli, with the displacements varying from 0 to 100 pixels. The results show that the cIOU behaves nearly the same as a IOU, which shows that the cIOU is a validated overlap metrics for detection of glomeruli.

\subsection{Rotation Consistency}
\begin{figure}
\begin{center}
\includegraphics[width=1\linewidth]{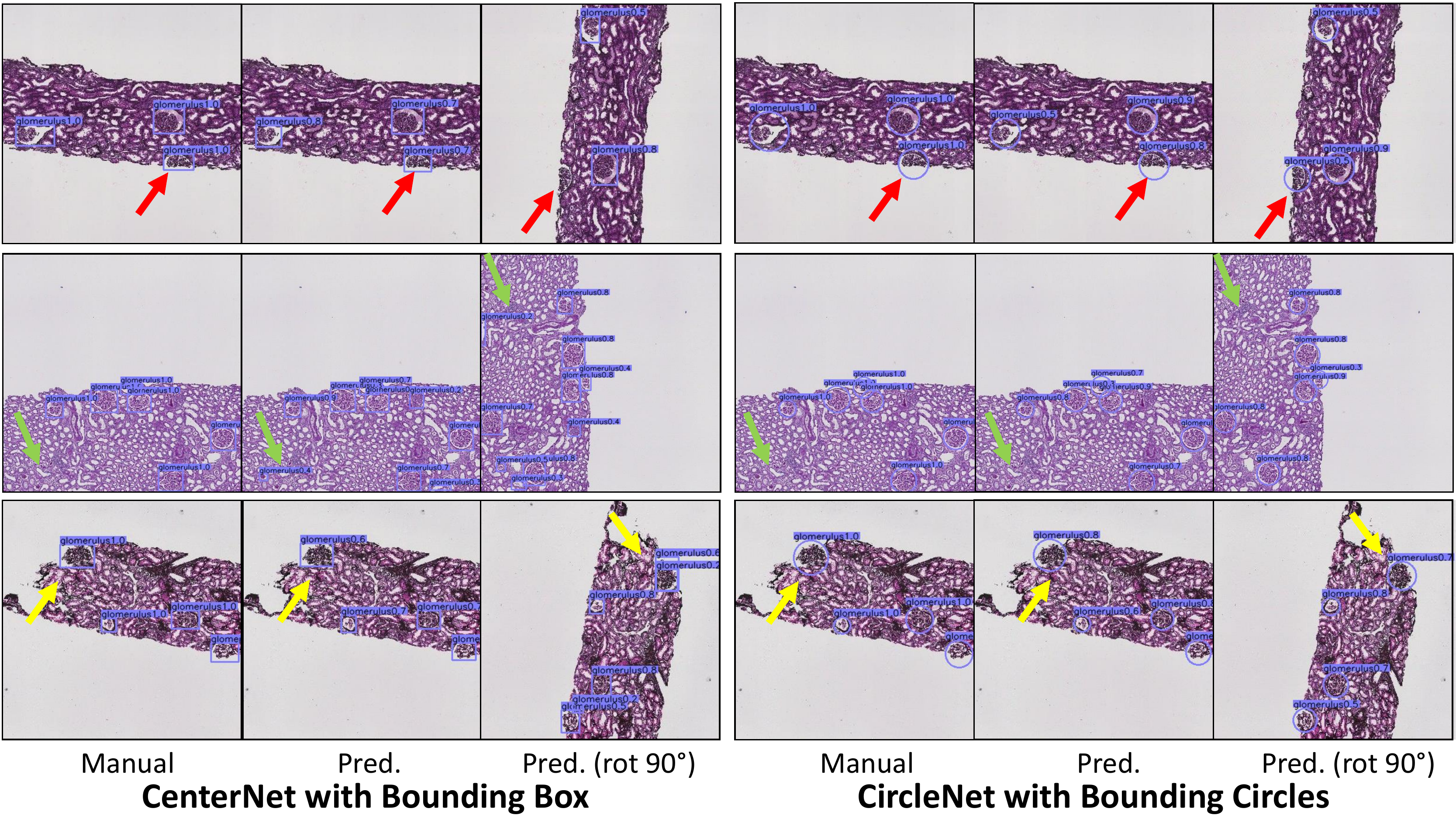}
\vspace{-2em}
\end{center}
   \caption{\textbf{The qualitative results of detection and rotation consistency.} Each row indicates a different example patches. The arrows indicate the inconsistent detection results in different cases.}
\label{fig:rotate}
\end{figure}
 
Another potential benefit for circle representation is better rotation consistency than rectangular boxes. As shown in Fig. \ref{fig:rotate}, we evaluated the consistency of bounding box/circle detection by rotating the original testing images. To avoid the impact from intensity interpolation, we rotated the original image 90 degrees rather than an arbitrary degree. By doing this, detected bounding box/circle on rotated images were conveniently able to be converted to the original space. To fairly compare the rotation consistency, the rotation was only applied during testing stage. Therefore, we did not apply any rotation as data augmentation during training for all methods. The consistency was calculated by dividing the number of overlapped bounding boxes/circles (IOU or cIOU $>$ 0.5 before and after rotation) by the average number of total detected bounding boxes/circles (before and after rotation). The percentage of overlapped detection was named ``rotation consistency" ratio, where 0 means no overlapped boxes/circles after rotation and 1 means all boxes/circles overlapped. Table \ref{table:overlap} shows the rotation consistency results with traditional bounding box representation and bounding circle representation. The proposed CircleNet-DLA approach achieved the best rotation consistency. One explanation of the better performance would be that the radii are naturally spatial invariant metrics, while the length and width metrics are sensitive to rotation.

\begin{table}
\caption{Rotation consistency results of bounding box and bounding circle.}\label{tab1}
\centering
\begin{tabular}{l@{}c@{\ \ }ccccccc}
\toprule
Representation & Methods $\quad\quad\quad\quad\quad$ & Backbone & Rotation Consistency \\
\midrule
Bounding Box & CenterNet-HG\cite{zhou2019objects} &  Hourglass-104 & 0.833 \\
Bounding Box & CenterNet-DLA\cite{zhou2019objects} & DLA & 0.851 \\
\midrule
Bounding Circle $\quad$ & CircleNet-HG (Ours) & Hourglass-104  & 0.875 \\ 
Bounding Circle $\quad$ & CircleNet-DLA (Ours) & DLA & \textbf{0.886}  \\ 
\bottomrule
\label{table:overlap}
\end{tabular}
\vspace{-2em}
\end{table}

\section{Conclusion}

In this paper, we introduce CircleNet, an anchor-free method for detection of glomeruli. The proposed detection method is optimized for ball-shaped biomedical objects, with the superior detection performance and better rotation consistency. The new circle representation as well as the cIOU evaluation metric were comprehensively evaluated in this study. The results show that the circle representation does not sacrifice the effectiveness with less \ac{DoF} compared with traditional representation for detection of glomeruli. 

\section{Acknowledgement}
This work was supported by NIH NIDDK DK56942(ABF), NSF CAREER 1452485 (Landman).


%
%
\clearpage
\bibliographystyle{splncs04}
\bibliography{main}
%




\end{document}